\documentclass{article}
\usepackage{preamble}

\title{

Compression ensembles quantify aesthetic complexity and the evolution of visual art
\vspace{-0.1cm}
}

\author{
Andres Karjus\textsuperscript{1}
Mar Canet Solà\textsuperscript{1}
Tillmann Ohm\textsuperscript{1}
Sebastian E. Ahnert*\textsuperscript{2,3}
Maximilian Schich*\textsuperscript{1} \\
\small \textsuperscript{1}ERA Chair for Cultural Data Analytics, Tallinn University  \hspace{1mm}
\small \textsuperscript{2}University of Cambridge \hspace{1mm}  
\textsuperscript{3}The Alan Turing Institute \\
\small *These authors contributed equally to this work as last authors.

}
\date{\small May 19, 2022 \vspace{-0.4cm}} %
\begin{document}
\maketitle

\begin{abstract}
    The quantification of visual aesthetics and complexity have a long history, the latter previously operationalized via the application of compression algorithms.
    Here we generalize and extend the compression approach beyond simple complexity measures
    to quantify algorithmic distance in historical and contemporary visual media.
    The proposed ``ensemble" %
    approach works by compressing a large number of transformed versions of a given input image, resulting in a vector of associated compression ratios.
    This approach is more efficient than other compression-based algorithmic distances, and is particularly suited for the quantitative analysis of visual artifacts, because human creative processes can be understood as algorithms in the broadest sense. Unlike comparable image embedding methods using machine learning, our approach is fully explainable through the transformations.
    We demonstrate that the method is cognitively plausible and fit for purpose by evaluating it against human complexity judgments, and on automated detection tasks of authorship and style.
    We show how the approach can be used to reveal and quantify trends in art historical data, both on the scale of centuries and in rapidly evolving contemporary NFT art markets. We further quantify temporal resemblance to disambiguate artists outside the documented mainstream from those who are deeply embedded in Zeitgeist.
    Finally, we note that compression ensembles constitute 
    a quantitative representation of the concept of visual family resemblance, as distinct sets of dimensions correspond to shared visual characteristics otherwise hard to pin down. 
    Our approach provides a new perspective for the study of visual art, algorithmic image analysis, and quantitative aesthetics more generally. 
\end{abstract}

\begin{figure}[htb!]
	\noindent
	\includegraphics[width=\columnwidth]{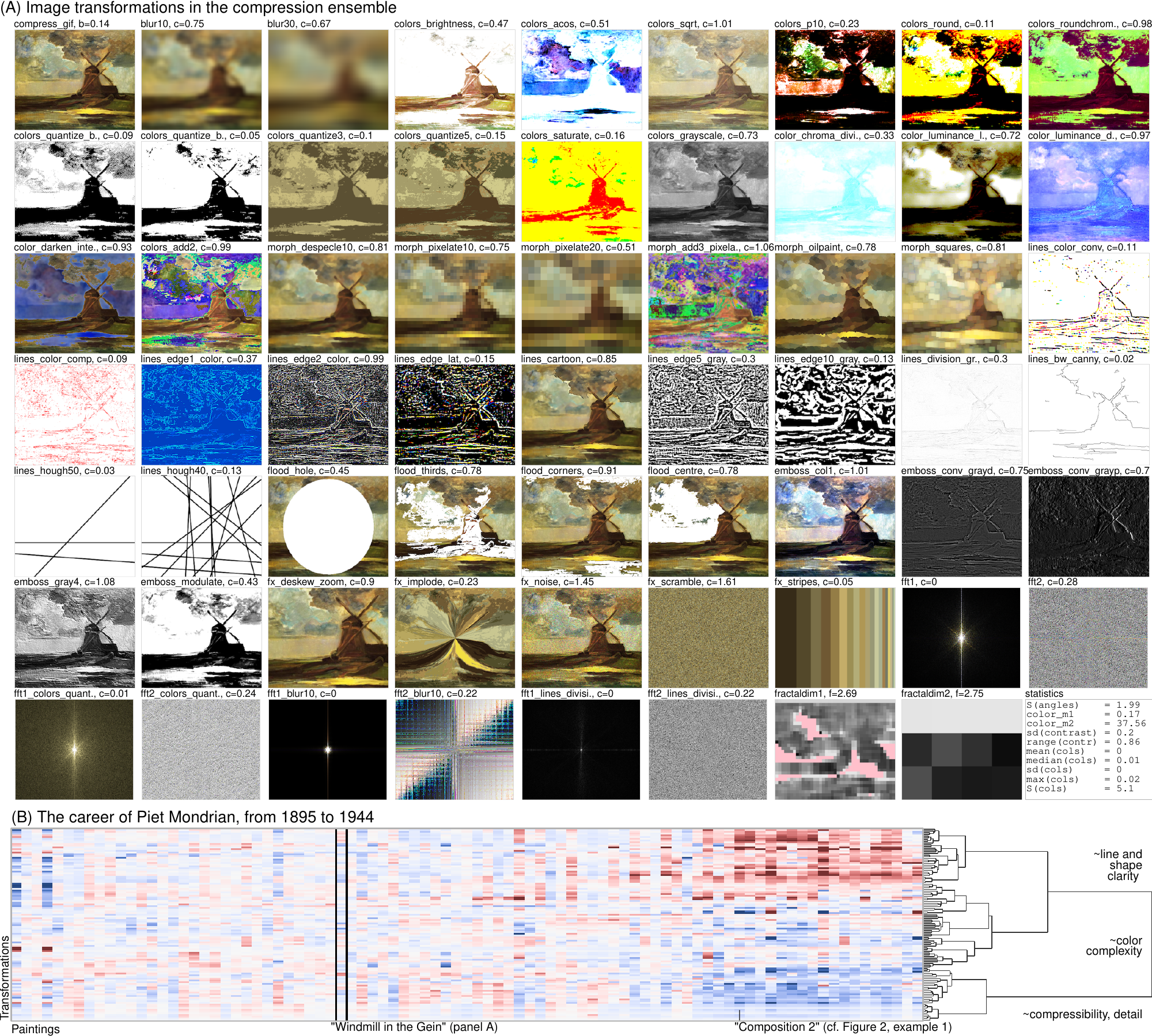}
	\caption{
	The compression ensemble of multiple image transformations allows for meaningful quantitative comparison of artworks. (A) Compression ensemble transforms for the example of ``Windmill in the Gein" (1906-1907) by Piet Mondrian. (B) Matrix of ensemble values for the oeuvre of Mondrian as recorded in our Historical dataset. The top left in (A) is the original image, with the compression ratio $b$ indicating the \textsc{gif}-to-bitmap ratio, i.e. the baseline compression size. $c$ values indicate the compression ratio against this non-transformed \textsc{gif} size. Turning this colorful painting into gray scale, for example, reduces its complexity and thus increases compressibility --- but would not affect an already gray scale pencil drawing. The ensemble also includes fractal dimension ($f$) and estimates of colorfulness.
	The brightness of the lightest and darkest example images is slightly adjusted here to make them perceptible.
	The complete ensemble further includes a subset of these transforms using a reduced size base image (see Materials and Methods).
	Each row in (B) represents a transformation,
	arranged by similarity; each column is an artwork. The Windmill example of (A) is highlighted with vertical lines in (B). The matrix values are z-scores, calculated using the mean and standard deviation of the artist's era (including all their own and contemporary works in our dataset). Darker blues indicate lower, and reds higher values compared to the respective average. Mondrian starts out fairly traditional, 1895 left to 1944 right, but eventually develops his iconic style, departing from the mainstream (see Example 1 in Figure \ref{fig_bigpca}B).  
	}\label{fig_illustrate}
\end{figure}
\section{Introduction}

The quantification of visual aesthetics, including artistic expression goes back to
\textcite{birkhoff_aesthetic_1933} and \textcite{bense_einfuhrung_1969}, inspiring several computational approaches in the recent past \parencite[cf.][]{ galanter_what_2003,rigau_conceptualizing_2007,kim_large-scale_2014,elgammal_quantifying_2015,sigaki_history_2018,elgammal_shape_2018,zanette_quantifying_2018,muller_compression_2018,lee_dissecting_2020}.
Previous research drawing on information theory has shown repeatedly, often in parallel, 
that subjective visual complexity can be estimated with some accuracy using compression algorithms such as \textsc{zip} or \textsc{gif} \parencite[][]{fairbairn_measuring_2006,rigau_conceptualizing_2007,campana_compression-based_2010,forsythe_predicting_2011,palumbo_examining_2014,guha_image_2014,chamorro-posada_simple_2016,machado_computerized_2015,muller_compression_2018,fernandez-lozano_visual_2019,ovalle-fresa_standardized_2020,bagrov_multiscale_2020,mccormack_complexity_2022,murphy_distributed_2022}. 
While some of the aforementioned proposals also included testing against perceptual human judgments, results diverge as to which single compression algorithm or approach would be optimal. Elsewhere in the humanities and cultural sciences, measures of compression length, taken at face value, have been used to compare the complexity of various visual inputs \parencite[e.g.][]{tamariz_culture_2015,miton_graphic_2021,han_chinese_2021}.
Closely related entropy-based information-theoretic approaches have also been applied to quantify artistic styles and conceptual groupings \parencite[][]{sigaki_history_2018,lee_dissecting_2020,tran_entropy_2021}.

When considering the quantitative analysis of visual art, it makes sense to adopt an algorithmic approach, since the creative processes of creating an artwork also follow a set of procedures --- or algorithms, in the broadest sense --- which can be assumed to be particular to a given artist and career period \parencite[cf.][]{bense_einfuhrung_1969}. 
Algorithmic complexity is best understood through the lens of algorithmic information theory \parencite{kolmogorov_logical_1968,chaitin_algorithmic_1977} which defines the complexity of a dataset in terms of the shortest algorithm that reproduces the data. 
While Kolmogorov complexity itself is uncomputable, the size of a compression of a given dataset can serve as its upper bound, and be extended to measures of algorithmic distance.
An established example for such a measure is normalized compression distance \parencite[NCD;][]{li_similarity_2004}, which however requires a separate compression for each comparison event \parencite[and has been rarely applied to visual materials; but see][]{cilibrasi_clustering_2005}. %
Pairwise image comparison frameworks have been also proposed by \textcite[][]{guha_image_2014,muller_compression_2018}.

Here we introduce a simple and fast algorithmic comparison framework for images, and apply it to the exploration of two-dimensional art such as paintings and drawings. In this ``ensemble" approach, an array of image processing filters is applied to each input, including various low and high pass filters, distortions and color manipulations (see Figure \ref{fig_illustrate}.A). The altered images are all compressed, yielding vectors of compression lengths (as a ratio, being divided by the size of the compression of the original bitmap image). This approach is augmented by another array of statistical transformations such as colorfulness metrics and fractal dimension. 
The latter could also be viewed as compressions in a broader sense; if the purpose of a given ensemble is to be strictly an estimate of Kolmogorov complexity via compression only, then these can be omitted of course. %
The resulting vectors can be rapidly compared, clustered, and used in downstream tasks such as identification of authorship or style, as demonstrated below. Fitting a new image into an already generated model does not require any retraining or realignment; just the set of applied transformations needs to match.

We use lossless \textsc{gif} to compress the transformed images, and additionally \textsc{png} and lossy \textsc{jpeg} on a smaller subset. In total, the current model consists of 112 transformations.
The exact number is unimportant and a question of optimization, as more (non-collinear) features provide more information but increase computation time, while different types of transforms are informative for different tasks. Indeed, as demonstrated in the art classification experiments below, a handful of well-chosen features can in some cases yield accuracy close to using the full ensemble. For a more detailed description of the workflow pipeline see the Methods and Materials section, and the Supplementary Information for the full list of transformations.

Some of the outputs of the transformations, e.g. blurs of various magnitudes, may well correlate. In applications where multicollinearity would be an issue, or where interpretation in a lower number of dimensions is desired, methods such as Principal Component Analysis (PCA) or UMAP can easily be applied. 
Here we use both, as the PCA is directly interpretable due to its linear relationship with original variables, while UMAP arguably provides better low-dimensional representations \parencite[cf.][]{mcinnes_umap_2018}.
The dimensions of the vector space, preceding PCA or UMAP, remain readily interpretable, as each represents a distinct transformation (see Figures \ref{fig_illustrate}.B and \ref{fig_sup_smallmaps}). The dimensions of a PCA also remain interpretable, due to the linear relationship between components and original variables. 

UMAP constitutes a complementary ``field of similarity" %
\parencite[cf.][]{riedl_structures_2019} where similar images cluster together intuitively, like in a spring-embedded network-diagram,
while remaining subject to the so-called "curse of dimensionality".
Yet, as shown in Figure \ref{fig_illustrate}.B (and \ref{fig_sup_smallmaps} in the Supplementary Material), we can break the curse by mapping the contribution of individual transformation or PCA dimensions onto the general UMAP, which effectively assumes the function of a reference topography. The resulting ``small multiple" visualization provides an intuition why and ensemble of multiple different transformations and compressions is necessary towards a fuller understanding of visual aesthetic complexity.

\begin{figure}[htbp]
	\noindent
	\includegraphics[width=\columnwidth]{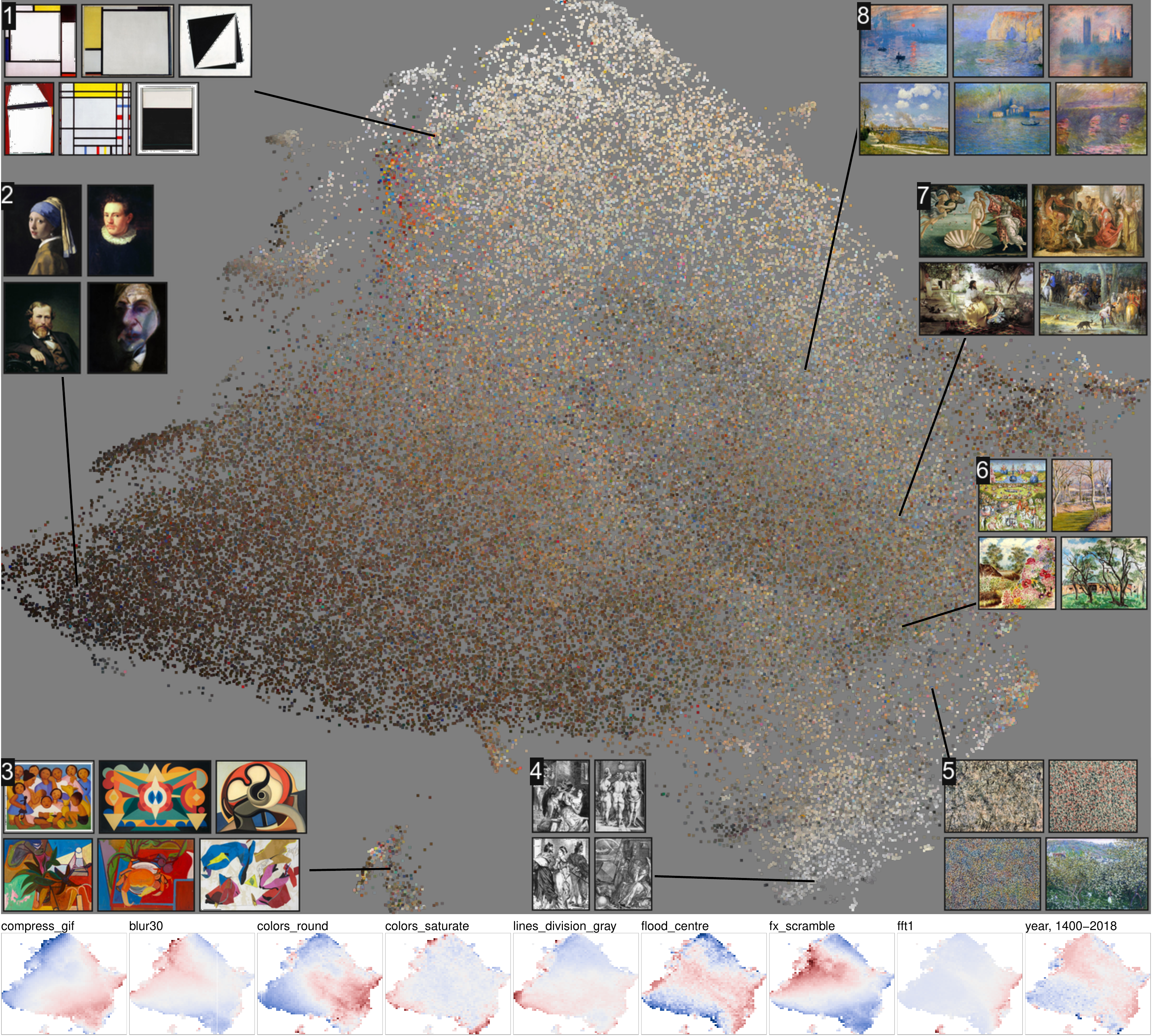}
	\caption{
	UMAP projection of the full compression ensemble space, constituted via 112 transformation variables, for 74k artworks in our Historical dataset 1400-2018. Each dot is an artwork, here reduced to a single pixel. 
	A number of examples are highlighted along with their cosine-nearest neighbors in the model (Examples 1-8).
	Proximity in this space indicates multidimensional similarity in aesthetic complexity, i.e. often by proxy, style or more general family resemblance. Images with few colors and simple structure are close together, and distant from those with complex patterns and palettes (Examples 1 versus 5). 
	Images that are close by often also contain similar subjects and color palettes, due to conventional commonalities in the aesthetics of depicting certain scenes and objects (cf. Examples 2 and 8). %
	The small bottom panels depict the same UMAP, yet as heatmaps colored according to the mean values of individual transformations in a given UMAP region, blue to red, low to high (cf. Figure \ref{fig_illustrate}, and \ref{fig_sup_smallmaps} in the Supplementary Materials for a full set of these maps). 
	While the 
	nearest neighbors sets pictured above intuitively make sense, the additional heatmaps strikingly clarify the underlying polymorphic complexity, promising a rewarding territory for future research.
	}\label{fig_bigpca}
\end{figure}

The nature of a transformation --- edge detection, color quantization, blurring --- emphasizes or attenuates particular characteristics of the input images.
For example, applying a black-and-white transformation to a colorful image increases its compressibility relative to the original (yielding a compression ratio $\ll 1$), but has no effect on a already black and white image (a compression ratio $\approxeq 1$). Applying coarse 
pixelation  %
to Piet Mondrian's geometric abstract paintings (Example 1 in Figure \ref{fig_bigpca}) barely changes their compressibility, whereas the same pixelation greatly increases the compressibility of highly detailed works, such as those of Hieronymus Bosch (Example 6 in Figure \ref{fig_bigpca}). A difference between two images in terms of their compression ratios for a given transformation reveals that they differ in this aspect.
Therefore, images similar in multiple aspects of aesthetic complexity end up close together in the multidimensional ensemble space, while dissimilar ones are placed far apart.

Similarity in the sense of depicted objects or scenes is not directly encoded, as the compression ensemble does not include any 
visual similarity features in the machine learning sense.    %
However, certain themes may be more popular than others within a given region of the space, and depicting certain things (such as people on a dark background) may yield a similar complexity profile, which is why nearby artworks often also contain similar subjects and color palettes (e.g. Example 2).

The multidimensional ensemble space of compression ratios, consisting of continuous values, also allows for interesting mathematical vector operations \parencite[not unlike in word embeddings, cf.][]{vylomova_take_2016} and explainable latent space exploration. As an example, adding the vectors of Example 4 and 6 of Figure \ref{fig_bigpca}, i.e. an etching or woodcut print plus the ``Garden of earthly delights", yields a vector where the closest neighbors are prints of trees and nature. Multiplying the Mondrian vector of Example 1 with the averaged vector of all landscape paintings nets an abstract landscape. See the Supplementary Materials for the results of summing the vectors of all pairs of examples called out in Figure \ref{fig_bigpca}.

Importantly and in contrast to previous research, our goal is not to construct a model to learn and predict what humans may perceive as visually complex or intuitively ``aesthetic" as such \parencite[cf.][]{forsythe_predicting_2011,cela-conde_sex-related_2009,fernandez-lozano_visual_2019}. Nor is it our goal to compare artworks based on the similarity of their depicted subjects, recognizable features, or iconographic attributes \parencite[cf.][]{tan_ceci_2016,mao_deepart_2017,elgammal_shape_2018}.
Rather, our model is meant to capture the residual signal of the generating process, in a kind of ``algorithmic fingerprint" of an artwork, to eventually quantify and explore the artistic dynamics and evolution in the space of intrinsic aesthetic complexity. However, to do so with confidence, we first verify our model in a number of experiments, including ones that use human judgment scores.

We then go on to quantify global trends in historical art over the past six centuries in a benchmark dataset, and over the course of the first 175 days of the non-fungible token (NFT) art marketplace Hic et Nunc. Finally, again on historical timescales, we introduce a temporal resemblance model to 
quantify artistic career trajectories, grouping them into qualitatively distinct types. 
We reveal artists that were well embedded in the historical tradition of their time, those who simultaneously experimented with different styles, artists with transitory success, and those who were later seen as ahead of their time.

\FloatBarrier
\section{Results}

We make use of two large art corpora to proof the application of the compression ensemble approach for visual data, while exemplifying the exploration of historical and contemporary dynamics of visual art.
The first dataset which we denote as ``Historical" (henceforth capitalized when being referred to) is illustrated in Figure \ref{fig_bigpca}. It is sourced from the art500k project \parencite[][]{mao_deepart_2017}, filtered to only include two-dimensional art with intact metadata (in particular, retrievable year of creation). Our subset contains 74028 (primarily Western) artworks representing 6555 artists. We note that after our filtering, the remaining dataset consists mostly of items art500k had in turn sourced from Wikiart.org. The latter is an online, user-editable, encyclopedic collection of mostly Western art images, which is also frequently used in computer vision research. 
From an art historical standpoint the dataset provides a reasonable and sufficient proxy benchmark to show the feasibility of our approach. Known biases of the Historical dataset include reliance on partially dated literature, including a corresponding gap of 18th century art, and very likely some variation in terms of reproduction quality due to the broad variety of the crowdsourced images, either found in the public domain or taken from a great variety of literature and online sources on the basis of fair use. 
Digitizing larger amounts of visual cultural heritage in high resolution, consistent quality, and minimal bias is a generational challenge. While the Historical dataset is sufficient for our proof of concept, as more data in better quality becomes available, descriptions based on our method are expected to also become more precise and representative.

The second dataset which we denote as ``Contemporary" is mined from Hic et Nunc, a Tezos blockchain-based NFT art marketplace, representing the first 175 days of its existence (March to August 2021; 51640 artworks, 7284 artists). It contains 31\% of all the objects added to the marketplace during our observation period of 175 days: we only include static images (\textsc{jpeg, png}) as we do not yet have a pipeline to compress multi-frame objects such as animated \textsc{gif}s and videos), exclude very small resolution images (such as icons), and a subset for which the data collection process failed to retrieve the image. %
For an overview of the NFT-driven ``crypto art" market, see \textcite[][]{nadini_mapping_2021,vasan_quantifying_2022}.

Before applying the compression ensemble approach to capture systematic patterns of art history, we evaluate it extensively using three datasets and two methodologies, (1) examining correlations of our model predictions with human judgments of visual complexity, (2) using the model to perform authorship and style attribution. We show that our model performs very well on the first task and with fair accuracy on the second task (despite not being trained for the specific purpose).
The second experiment also demonstrates explicit connections between specific dimensions in the vector space of compression ratios and particular aspects of the corresponding artworks. For example, the compression ratio of edge-filter transformations are informative regarding the genre of the work (portraiture versus landscape), while color-affecting transforms can help predict the medium (drawing vs oil painting).

\subsection{Evaluation} 
\subsubsection{Human complexity norms}
\begin{figure}[htb]
	\noindent
	\includegraphics[width=\columnwidth]{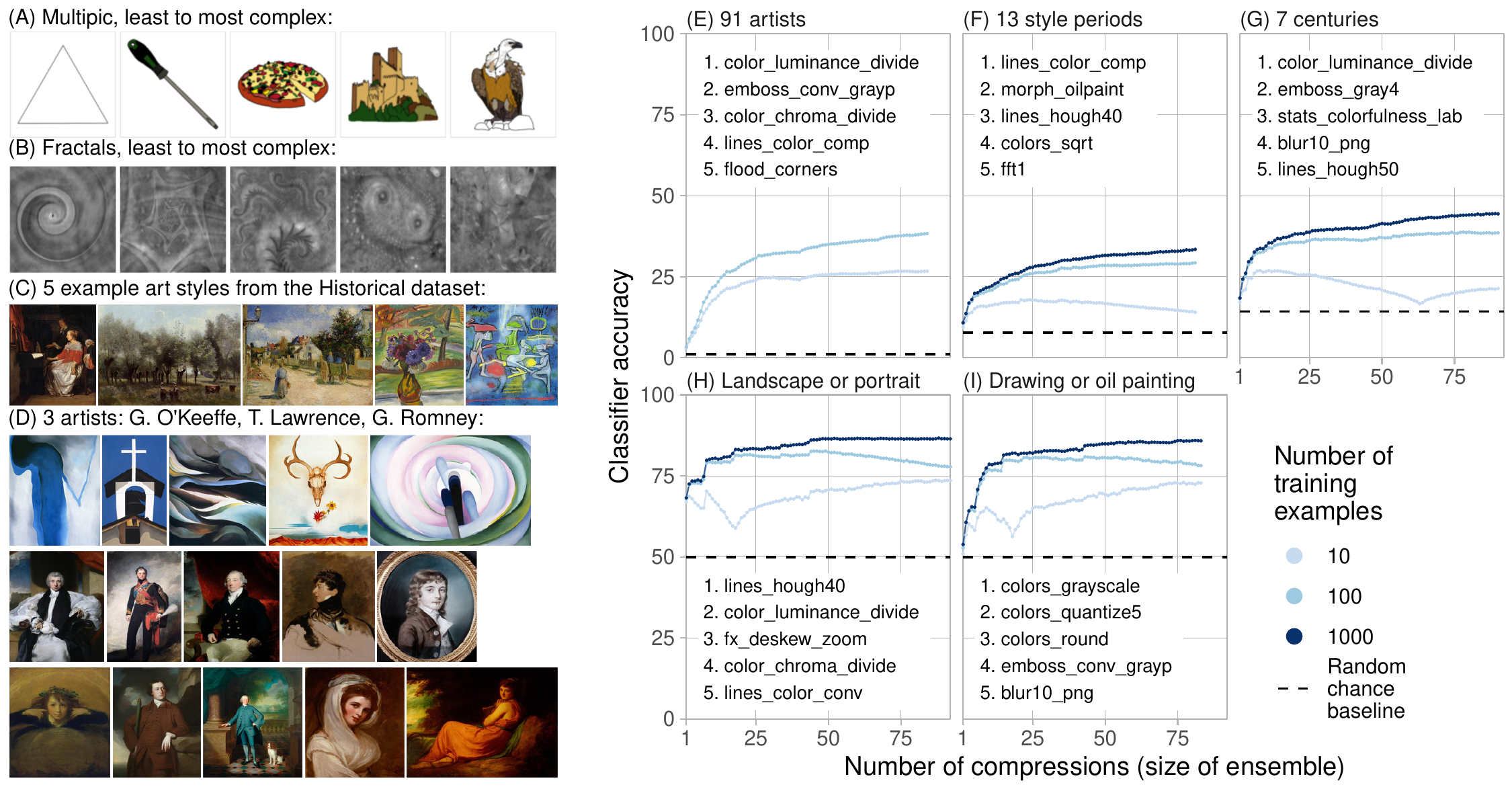}
	\caption{
	    The compression ensemble approach is cognitively plausible and also performs well in algorithmic prediction of artwork authorship, date, style, genre, and medium. 
	    (A) and (B) exemplify the two human ratings datasets, Multipic and Fractals (evaluation: see text). The five art style period examples (C) represent Baroque, Realism, Impressionism, Expressionism and Surrealism via central images in the ensemble for each style. Panel (D) illustrates the difficulty of the artist detection task: while some artists are more unique and hence recognizable (O'Keefe), others produce very similar works, while also changing over their careers (Lawrence, Romney).
	    Panels (E-I) illustrate mean testing accuracy given variable number of training items (light to dark blue) and number of transformations used (horizontal axis; the total number of features varies between tasks, as zero-variance and collinear ones are excluded). The dashed horizontal line indicates baseline chance accuracy for each task. Each dot stands for one added transformation feature, always starting with \textsc{gif} compression without transformation. The next 5 are given on each panel. Different transformations, ordered by variable importance, are informative in different tasks, e.g. color-related transformations in distinguishing paintings from drawings. Just compressing the image without transforming already provides an above-chance result in all cases, even if using just a handful of training examples. Adding more transformations generally improves performance (when there are enough training examples to avoid overfitting; dark blue dots). That being said, around 15-20 well-chosen features are usually already enough to get close to maximal performance.
	}\label{fig_evals}
\end{figure}

We assess the cognitive plausibility of the compression ensemble approach by comparing its predictions of visual complexity with human judgment norms from two datasets. The first dataset, MultiPic \parencite[][]{dunabeitia_multipic_2018} consists of 750 colored pictures of concrete concepts, and human judgments on various aspects of visual perception, including complexity, based on experiments with a total of 620 participants from six language communities (British English, Spanish, French, Dutch, Italian, German; see Figure \ref{fig_evals}.A). The dataset does not include individual ratings, only means for each image for a given language sample. The second dataset, Fractals \parencite[][]{ovalle-fresa_standardized_2021} consists of 400 abstract fractals and related norms, again including (means of) judgments of visual complexity, here by 512 German-speaking participants (Figure \ref{fig_evals}.B).
Previous research has also engaged in analogous exercises of evaluations against human complexity judgements \parencite[][]{machado_computerized_2015,mccormack_complexity_2022}. We use the datasets described here as they are both publicly available while representing fairly large pools of participants. %

We generate the compression ensemble vectors separately for each of the two datasets, then carry out repeated out-of-sample evaluation where we train a linear regression model on a set of vectors to predict human scores, then test its accuracy on a separate test set. The results are very good, with median absolute error ranging from $0.19$ (Multipic English) to $0.23$ (Multipic Flemish) on a scale of 0 to 5. To put this in perspective, this is smaller than the differences between languages in this dataset (the median standard deviation of complexity scores per image across languages is $0.24$). In Fractals, median absolute error is $0.46$ on the same scale of 0 to 5. The linear regression model with compression ratios as predictors describes most of the variance (measured as adjusted $R^2$) in human visual complexity ratings: $73\%$ (Multipic Italian) to $83\%$ (Multipic Flemish), and $32\%$ in Fractals. By comparison, using \textsc{gif} compression alone describes just $37-44\%$ (Multipic) and $10\%$ (Fractals).
These results provide us with confidence that the approach is cognitively valid, correlating with what the human eye would consider visually complex.

\subsubsection{Artist, date, style, genre, and medium classification}

The second evaluation involves the Historical dataset, in the form of a number of retrieval or classification experiments. We generate the compression ensemble vectors for the entire dataset, and extract the following subsets, where each included class has at least 1100 unique examples:
13 style periods as per metadata (5 of which are exemplified in see Figure \ref{fig_evals}.C),
7 centuries,
drawings vs oil paintings,
landscape paintings vs human portraits,
and 91 artists with at least 110 artworks each.

We perform out-of-sample evaluation where we repeatedly train a classifier for each subset, on a randomly sampled set of vectors from each class in the subset to predict the relevant class labels such as style period (n=1000 per class, except 100 for authors due to limited data), then test its accuracy on a separate test set (n=100 per class, except n=10 per author). We use Linear Discriminant Analysis --- a simple, computationally lightweight supervised machine learning model that straightforwardly generalizes to multi-label classification. To probe how well the ensembles work on this task given different amounts of data and number of transforms, we carry this out in a step-wise manner, as depicted in Figure \ref{fig_evals}.E-I. Each classifier is trained on 10, 100 and 1000 examples of each class, and employing an increasing number of transforms, starting from the baseline of \textsc{gif} compression (ratio to raw bitmap file size). The rest of the features are ordered by a rough estimate of variable importance (derived from repeatedly training binomial logistic regression classifiers on all possible pairs of classes and averaging the t-statistics of the variables). %

Even with a handful of examples and a couple of the most informative transforms, the simple classifier is able to detect above chance the creator, the date, style, genre, and medium of a given artwork. With a 100 examples and the full ensemble of transforms, author (n=91) detection accuracy is 38\%, which is much higher than the accuracy of 1.1\% that random attribution would achieve by chance.
Provided 1000 examples per class, oil paintings are distinguished from drawings about 86\% of the time, same for landscapes vs human portraits (both have 50\% random chance baseline), style period 34\% (baseline $\sim8\%$) and century 44\% (baseline $\sim14\%$).

The ranking of the transforms beyond the compression baseline (as depicted in Figure \ref{fig_evals}.E-I) is also informative. The aspects represented by the transformations vary in usefulness in the prediction task; for example, gray-scaling distinguishes pencil drawings from colorful oil paintings, because this is one of the primary aspects they differ in. Turning this around, the explainable features of the compression ensemble can be used to describe how any two images (or sets of images) differ, by looking into which transformation dimensions describe the most variance.

Inspecting the relevant confusion matrices reveals the errors are fairly systematic and intuitive, as classification errors are more likely between adjacent style periods and artists (see Figure \ref{fig_sup_stylecm} in the Supplementary Materials). In the set of 91 artists, Thomas Lawrence and George Romney are most often confused with each other by the model --- and indeed, both are portrait artists from roughly the same period (see Figure \ref{fig_evals}.D). Conversely, artists in a distinguishable style or genre are easy to identify, for example the 19th century engraver Charles Turner is detected at 97\%. Rococo, also know as Late Baroque, is correctly labeled in 47\% of the tests, while 16\% of it is misclassified as Baroque. Impressionism is easiest to identify (53\% correct) --- but confused with Post-Impressionism (14\%). Expressionism is by far the hardest to put a finger on (12\%).

Good results in authorship and style attribution have also been achieved using purpose-trained classifiers built upon large pre-trained deep learning models \parencite[cf.][]{mao_deepart_2017,strezoski_omniart_2017,elgammal_shape_2018}. For reference, although not directly comparable due to training and test set differences, \textcite[][]{mao_deepart_2017} report a 39\% accuracy for style period and 30\% for author retrieval; \textcite[][]{tan_ceci_2016} report 55\% for style and 76\% for artist (but that is between just 23 artists with the most training data). If for example authorship attribution was the goal, we envision that the accuracy of such models could likely be improved further by combining them with compression ensembles. 
The purpose of this exercise here however is not to compete with these approaches, but to show that a compression ensemble --- despite consisting of no features other than file size ratios and statistical transformations, and containing no pre-trained baseline --- still captures and disambiguates enough family resemblance to place stylistically similar artworks close together and dissimilar ones apart, with a non-random error structure.

\FloatBarrier
\begin{figure}[htbp!]
	\noindent
	\includegraphics[width=\columnwidth]{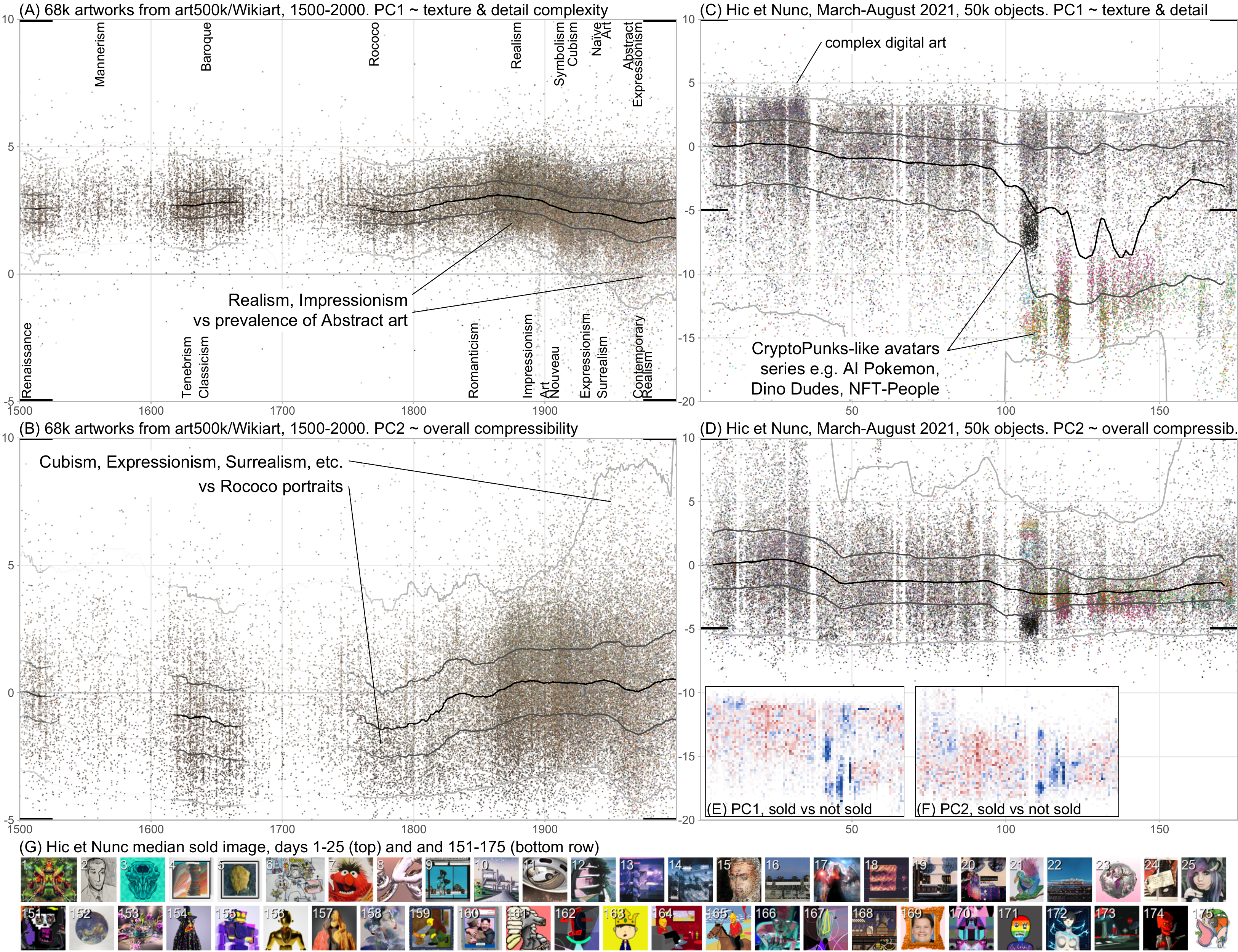}
	\caption{
	Aesthetic dynamics over 500 years in the Historical dataset 1500 to 2000 (left, A \& B), and over the first 175 days of the contemporary NFT art market Hic et Nunc from March 2021 (right and bottom, C \& D). Each dot is an artwork, reduced to a single pixel. The principal components (vertical axes) are based on a concatenation of both vector sets, making the graphs comparable. 
	(A) and (C) show the joint first component PC1, (B) and (D) the second PC2, also allowing for a reading across datasets, left to right. Note however the different ranges on the vertical axes: the Historical dataset is constrained to a much smaller area in the aesthetic complexity space (marked by black brackets on the sides).
	The trend lines correspond to the median (black) and quartiles (dark gray) of a given principal component; 95\% of the data lies between the outer light gray lines. 
	The more frequent style period labels are given in (A), arranged by the median year of the respective artworks.
	The insets (E) and (F) indicate areas of the complexity space conductive to NFT sales (dark red means all items in a given area were sold; dark blue that none was sold). 
	The bottom panel (G) shows typical NFTs sold on the Hic et Nunc marketplace, as images closest to the median (across all PCs) for each day. Various avatar or portrait series (similar to CryptoPunks or Bored Ape Yacht Club) eventually rise to be among the most commonly minted objects --- visible as tight colorful groupings at low complexity in PC1 --- but not all such series are successful, as indicated by the prevalent blue areas in the corresponding inset panels.
	This example demonstrates how the same method can be used to make sense of both very long and very short timescales, in art history and contemporary art. 
	}\label{fig_dynamics}
\end{figure}

\subsection{Tracking historical and contemporary art dynamics}

Given the explainable nature of the compression ensemble vectors, and their cognitive and technical plausibility as demonstrated above, we can now use this method to investigate and interpret aesthetic trends over time. We do this for both the Historical and the Contemporary NFT dataset.

To simplify this task, we apply PCA here and focus on the two first most informative principal components. We obtain compression vectors for both the Historical and the Contemporary datasets, and fit them both in the same PCA space for comparability. 
Figure \ref{fig_dynamics} depicts change over time in these two components. 
The trend lines are all estimates from a rolling window of $\pm10$ (years of Historical data, days of Contemporary Hic et Nunc). Where there is insufficient data, the window is stretched up to size 50 to include at least 1000 artworks where possible; these broader estimates are reflected by decreased line opacity. Here the Historical dataset is limited to 1500-2000, as both ends outside of that range are quite sparse.

Changes in the trends in the half-millennium dataset correspond broadly to art historical style classifications. PC1 in this model corresponds to texture and detail complexity (loading onto blur, despeckle filters, and the Canny edge transform). There is a marked decrease (visible in the right half of Figure \ref{fig_dynamics}.A) going from the period of more detailed paintings of Realism and Impressionism to the second half of the 20th century where (in this dimension on average less complex) styles such as Abstract Expressionism and Pop Art become more prevalent. 

PC2 corresponds to overall compressibility (loading onto compression of the original unfiltered image with an array of algorithms). The median in the Historical dataset is lower where the dataset contains many Rococo style portraits (in the middle of Figure \ref{fig_dynamics}.B), which typically contain plain and therefore easily compressible areas, not unlike the pixel-art portraits of Hic et Nunc,  (cf. days 100-150 in Figure \ref{fig_dynamics}.C-D).
The PC2 values in the Historical data (Figure \ref{fig_dynamics}.B) go up around the onset of Impressionism, and the bounds are pushed once more with Cubism, Expressionism, Surrealism, and the general diversification of classic modern ``-isms".
As demonstrated in the Evaluation section above, given a sufficient number of transformations, such differences are consistent and diverse enough to predict style periods with reasonable accuracy.

While Historical and Contemporary data is combined in the same space, the vertical axes representing the principal components in Figures \ref{fig_dynamics}.A-B versus C-D are intentionally different, as the two datasets occupy markedly different ranges in the complexity space, with much higher variance in the Contemporary Hic et Nunc dataset compared to the more conventional Historical dataset. This does not necessarily mean that art in the last 500 years has been less creative or explorative. The relative boundedness instead is more plausibly rooted in a combination of material affordances and limits of curation and scholarship. The latter is a function of cultural selection, as collectors, audiences, and art historians put a bound on what has been and is considered worthwhile of adding to collections from the time of creation to current retrospectives. 

In contrast, everybody who is able to pay the fairly low minting fee can upload an artwork to blockchain art market places such as Hic et Nunc, making their creations public in an attempt to get attention and sell. Material affordances can further explain changes within the Historical dataset, and salient differences in relation to Contemporary NFT art. The Historical broadening of the parameter space goes in lockstep with the fraction of noted creatives growing faster than world population in the last five centuries \parencite[][]{schich_network_2014}. It is broadly established knowledge in art history that new technologies and concepts, from pigments to theories of perception \parencite[cf.][]{gombrich_art_1960}, were harnessed by said creatives, arguably at an equal pace. %
Examples include the emergence of more affordable blue pigment alternatives to the rare and expensive azurite and lapis lazuli, or (color) photography, which put traditional pictorial conventions of depiction into question.
Another striking difference between the Historical and the Contemporary NFT dataset becomes visible in Figures \ref{fig_dynamics}.A-B versus C-D when we focus on the range of colors in the single-pixel reductions of the artworks. The digital NFT images appear darker and more saturated, as they are using the full RGB color space, while the dominant color of Historical artworks tends to remain in the range of "natural" pigments, which one could buy in a physical art supply store.

Since we have information on transactions in the Hic et Nunc dataset (as of the data collection time, 22 August 2021), successful sales are shown as inset heatmaps E and F in Figure \ref{fig_dynamics}.D. The heatmaps show the fraction of sales across the first and second principal components respectively.
About half the objects in the Hic et Nunc sample in total were sold off by their authors during our observation period, with some areas --- dark red in the insets --- being clearly more conductive to sales, while others do not sell at all. Even qualitatively, one can see revealing patterns, such as the mass-minting of initially non-selling NFT images starting around day 110 in mid 2021 (including CryptoPunks-like avatar series, such as ``AI Pokemon", ``Dino Dudes", and ``NFT-People"). An emergent quality of these mass-produced images is that their texture and detail complexity (PC1, Fig. \ref{fig_dynamics}C/E, day 100-150) is substantially lower than the all preceding art, putting them more in the realm of icons or brand logos. 
At the same time their overall compressibility (PC2, Fig. \ref{fig_dynamics}C/F) is not only systematically lower, but also subject to much less variance, indicating that most of them are indeed low effort attempts to make money quick. 
The narrowness of the mass-produced NFT series also expresses itself in their skew towards highly saturated primary dominant colors.
In at least one case, this indeed seemed to work, where sales follow in the wake of a strong minting burst, mostly consisting of the ``NFT-People" and ``NFT Kids" series (cf. the rightmost vertical blue line in Fig. \ref{fig_dynamics}.E, followed by a light red wake). 

Taking a quantitative perspective, we further trained another Linear Discriminant Analysis model on the sales data, predicting whether an NFT art piece was sold or not, by the values in the compression vectors. Using training sets of size 20k per class and separate test sets of 5k (and replicating the model 500 times), the model predicts sales at an average accuracy of 58\% (or a 17\% kappa, given the 50-50 baseline). This is despite containing no information on the prestige or reach of the artists, past sales, the depicted content, nor trends of the market of the respective time. 
A linear regression model fitted to 23370 sold items predicting log price (excluding zero-price giveaways) by all the compression variables describes about 6\% of variance (adjusted $R^2$); allowing for interaction with the time variable improves this to 8\%.
While these are all fairly low scores in absolute terms, we consider this a promising result for future research, as it could be likely improved by combining our model with the aforementioned variables of author properties, sales history and past trends \parencite[see also][]{lakhal_beauty_2020,vasan_quantifying_2022}, to predict future trends in evolving art markets.

\FloatBarrier

\subsection{Quantifying temporal resemblance in art}

We can also use compression ensembles to investigate how the oeuvres of individual artists are situated in their eras. 
Tracing "the lifes of the artists" has been a central direction in the historiography of art since the 1550 book by Giorgio Vasari which initiated the genre \parencite[][]{vasari_lives_1998}. Since then, a great number of artist monographs and critical catalogs have filled several large art libaries around the world. More recently, multidisciplinary science has tackled the issue using methods of network science and quantitative measures of success, with some limiting their focus on birth to death migration \parencite[][]{schich_network_2014}, or on contextual and socio-institutional aspects, such as exhibition records and art market price information \parencite[cf.][]{fraiberger_quantifying_2018}, while yet others have taken into account visual aesthetic aspects using information theory or deep machine learning \parencite[][]{lee_dissecting_2020,liu_understanding_2021}. Other related work has looked at the innovativeness of individual artworks using deep learning \parencite[][]{elgammal_quantifying_2015}.

\begin{figure}[thbp]
	\noindent
	\includegraphics[width=\columnwidth]{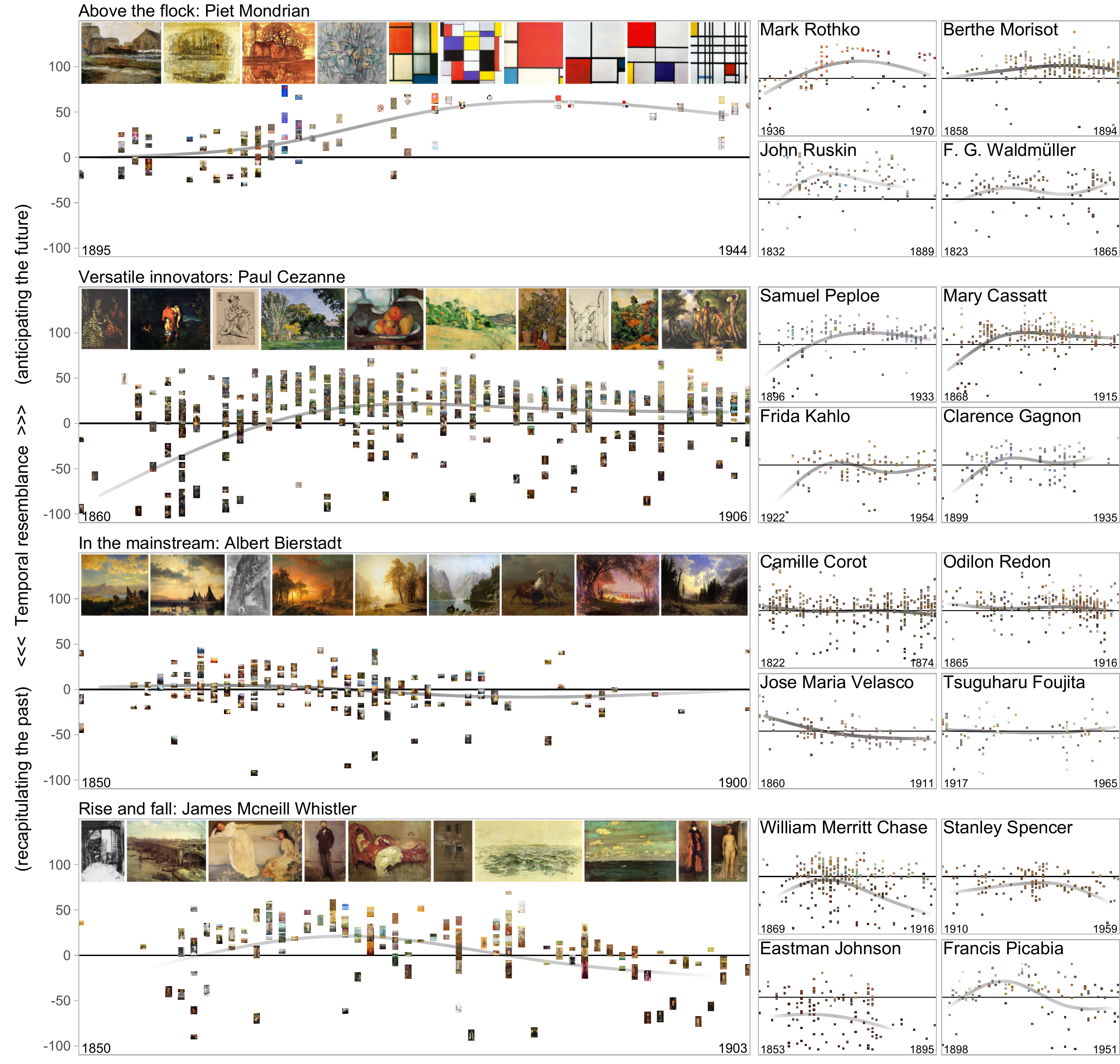}
	\caption{
	Several qualitatively different artist career types emerge from quantification through the lens of aesthetic complexity and applying the temporal resemblance model. Each small thumbnail is one work of art (see the Supplementary Materials for larger thumbnails). The vertical axis represents temporal resemblance. Points below zero correspond to artworks resembling artworks in the past. Points above zero anticipate art that has not yet been created, as their closest neighbors in compression space lie in the future.
	The curved lines are GAM fits; lighter shades of gray correspond to larger error, i.e. data not conforming to the curve. Examples over time close to the curve are displayed as a strip of larger thumbnails.
    	The larger panels on the left depict four arguably distinct examples of career trajectories, each associated with four smaller panels on the right, providing additional corresponding examples. We find artists that rise above the flock like Mondrian, versatile yet constant innovators like Cezanne, mainstream artists like Bierstadt, and artists that rise and fall relative to the mainstream like Whistler.
	}\label{fig_temporal}
\end{figure}

Here, we introduce a simple metric, which we call temporal resemblance, which goes beyond these approaches. Given that the vectors of all artworks reside in the same space (as set up above), we can calculate the nearest neighbors for each work. We use cosine similarity and the top closest 100 neighbors, while works by the same artist are excluded. The median of the temporal distances of these neighbors from the target work indicates if it resembles the past or anticipates some yet unseen future. This allows us to group artists who are traditionalist or historicist, those who stay current, and those ahead of their time. 
We also adjust the median time distances to account for the boundedness and density bias of the dataset: the metric reported here is derived from the residuals of a generalized additive regression model (GAM), still on the same yearly timescale (see Materials and Methods).
Figure \ref{fig_temporal} depicts the careers of 20 artists, grouped by career trend similarity.

Our metric is relative to the point in time of each work, and all measures are relative to all other works. Therefore, curves that stay close to the zero line in Figure \ref{fig_temporal} should be interpreted as artists who produce works that are similar to other artworks made in the same years, in terms of aesthetic complexity (and thus aspects of their style). 
That does not preclude changes in their style, if the changes in the artist and their era correlate. 
Staying around the zero may also indicate that a given artist is surrounded by a handful of prolific contemporaries with very similar output, who as a group may not be representative of the mainstream.
Descending curves can indicate an artist who becomes more traditional, the world catching up to an artist's style, or the world adopting other new styles. Note again however, while our results are intuitively correct for a trained art historian, the career comparisons discussed here only refer to artworks that are present and dated in our dataset.

Figure \ref{fig_temporal} reveals different modes of artistic existence, similar to yet not identical to the narrative types of Vonnegut \parencite[][]{reagan_emotional_2016}. Some 
such as Piet Mondrian or Mark Rothko, ``rise above the flock", starting out in the mainstream, but growing into their own distinct style, with works that could be considered ahead of their time (recall also Figure \ref{fig_illustrate}.B). 
Paul Cezanne and Mary Cassatt instead become ``constant innovators", starting out by producing more conventional, retrospective works, but growing and remaining innovative throughout the rest of his careers. 
Albert Bierstadt and Camille Corot represent ``mainstream artists", appearing more narrow in their practice, and remaining consistent with the current of their peers. 
Finally, we find artists who ``rise and fall", growing to their moment in history, then becoming more conventional again over the course of their careers. Examples would be James Mcneill Whistler or William Merritt Chase. An extreme case would be Eastman Johnson, who was predominantly drawing inspiration from the past, even at the height of his career.
That said, even highly innovative careers may also include such revivals on occasion. The oeuvre of Paul Cezanne, for example, contains some works resembling artworks preceding his own by 100 to 200 years.

As a static graph, Figure \ref{fig_temporal} is of course not comprehensive, but merely exemplifies how quantitative compression ensembles can be used to filter and cluster artistic trajectories from the multidimensional space of aesthetic complexity. 
We note that interactive versions of such plots could function as a research instrument for qualitative experts, to investigate the quantitative model and dataset biases, and compare artists between different datasets. The supporting material contains an alternative version  Figure \ref{fig_temporal} with larger overlapping thumbnails, allowing for a micro-macro-reading of qualitative content versus aesthetic quantification.

\FloatBarrier
\section{Discussion}

Products of human culture, such as art, language and music are all subject to ongoing 
change, complex dynamics, and cumulative evolution \parencite[][]{boyd_why_1996,tomasello_cultural_2009,beckner_language_2009-2,mesoudi_what_2018}. And even though complexity could emerge from a simple generating mechanism in principle, a single measure would likely prove insufficient to capture the polymorphic complexity of human cultural interaction and cultural products \parencite[cf. also][]{ebeling_komplexe_1998}. 
Here, we have demonstrated the utility of compression ensembles to quantify polymorphic visual aesthetic complexity, using fully explainable aspects in the process. 
We evaluated the cognitive plausibility of our approach, tested its viability at author, date, style, genre, and medium detection tasks, and showed how the approach can recover and reveal meaningful patterns in datasets of historical and contemporary art. Given the increasing availability of cultural datasets in machine-readable form, this operationalization opens up new avenues to study the dynamics of visual art at scale, over long time spans and almost in real time.
As such, the approach may help to transcend the still considerable specialization and bifurcation of qualitative art historical scholarship (by artist, period, region, style, genre, etc.). Revealing emergent patterns, while allowing for the study of systematic bias in comparison, indeed, our approach may fill a similar niche in art history as computational corpus linguistics does in relation to the qualitative study of literature. 
As each transformation in our compression ensemble represents a tangible visual aspect such as abundance of detail or colorfulness, as a whole, the ensemble constitutes a functional estimate of the philosophical and cognitive concept of polymorphic visual family resemblance, as originally used to characterize the similarity of games such as chess and soccer, later extended to polymorphic visual perception  \parencite{wittgenstein_philosophical_1953,weitz_role_1956,rosch_family_1975}. As shown in our evaluation experiments, our model captures enough polymorphic family resemblance to cluster similar styles and works by the same artist together. The explorations of the Historical art dataset yield results that meaningfully reflect the art historical scholarship which underlies the chosen dataset (Figures \ref{fig_dynamics}, \ref{fig_temporal}), yet here presented in easy to digest plots and visualizations. Figures \ref{fig_temporal} and \ref{fig_sup_temporal}, for example, summarize the careers of several artists in a single page, in each case reflecting the intuition of an individual connoisseur, who has been trained on the given corpus. We consider this a crucial contribution of our approach, as the recognition of visual family resemblance has hitherto remained particularly hard to explain. At the same time, the recognition of visual family resemblance is arguably foundational and intrinsically mastered by trained human art connoisseurs, by other human visual experts such as radiologists, and more recently by trained convolutional neural networks in deep machine learning \parencite[][]{lecun_deep_2015}. While the latter has solved the recognition of polymorphic family resemblance, including 
object detection,
which for a long time has remained computationally intractable, the magic of connoisseurship has so far still remained hidden in latent variables. It is in this sense a striking result that our evaluation results show the distinguishing explanatory power of taking into account multiple different explainable transformations for compression, effectively addressing what Friedländer in his foundational book on art connoisseurship called ``the visible in its manifoldness and unity, bristling against concepetual segmentation, so that the boundaries between the species of images get into flow" \parencite[][p. 60; our translation]{friedlander_von_1946}.

While our application of the methodology here has been aimed at visual aesthetic complexity, the same basic approach could be used to make sense of other, related phenomena.
For example, \textcite[][]{sinclair_beethoven_2022} raise the concept of ``aesthetic value", the ``attractiveness" of a given product of culture, 
to discuss whether the arts could be considered as a product of cumulative cultural evolution \parencite[cf.][]{mesoudi_what_2018}. 
They cast doubt onto the possibility of art or music objectively improving over time, which ``cumulative" would allude to \parencite[cf. also art historian][]{gombrich_ideas_1971}. However, we do not see an issue here, nor with their point that attractiveness is subjective to individual preference. A style that builds on or grows out of another style is not necessarily objectively better, but may better meet the preferences of its consumers in a given time, place, or ecological niche. This is not unlike the concept of communicative need in the context of language: a structure or lexical configuration may not be better in some absolute terms, but may be more optimal or efficient given the usage tendencies or needs of the language community \parencite[cf.][]{kemp_semantic_2018,karjus_conceptual_2021-1}.
The extent this can be studied depends on the data available. For example, the Historical dataset used in our work represents only a rough estimate of a (primarily Western, and somewhat dated) preference consensus, and even that with notable sampling caveats (as discussed above). The Hic et Nunc dataset is already more specific and also includes artist and collector profiles with trade and price information, which could be (carefully) interpreted as preference, and easily linked with the social media activity of the sellers and buyers for further study.
In this paper, we focused on static, two-dimensional art such as Historical paintings, drawings, prints, and Contemporary digital art in the case of the Hic et Nunc dataset. However, there is no intrinsic reason why the same methodology could not be applied to quantify other static media such as photographs, maps, websites or natural patterns \parencite[cf.][]{zanette_quantifying_2018,dou_webthetics_2019,fairbairn_measuring_2006,bagrov_multiscale_2020} to assess their aesthetic complexity (and by proxy, style) in a transparent, explainable framework.
Multi-frame visual media such as films and animations could be split up by frame or shot, and represented as sets of vectors in a compression ensemble.
Three-dimensional objects such as sculptures, architecture, or clothing items in fashion can similarly be operationalized by systematically scanning them from multiple angles, or using three-dimensional versions of the transformations and compressions, e.g. using voxels instead of pixels.
It may be also possible to employ this approach to quantify the aesthetic complexity of sound and music \parencite[cf.][]{beauvois_quantifying_2007,clemente_musical_2022} using the same method, by generating the spectrogram of a given sound, then applying the visual transformations on that.
Alternatively, the same compression ensemble principle could be applied directly to audio data, but using audio filters instead of image filters and compression of audio files in place of \textsc{gif} (analogously, visual filters directly on video files, or general filters on general signals). These avenues remain a prospect for future research for now.

There is also no reason why multiple ensembles or embeddings could not be concatenated in the case of multi-modal media, provided there is a principled way to weigh or normalize their contribution (the simplest way to do so would probably be PCA). As stated in the Introduction, our focus here was on aesthetic complexity and not visual similarity of recognizable subject features (such as faces, bodies, or objects), but the latter could easily be incorporated by horizontally aligning and concatenating our compression ensemble with a deep learning induced image embedding \parencite[e.g.][]{mao_deepart_2017}, or more explicitly by daisy-chaining feature recognition using deep learning and image segmentation, followed by compression ensembles of comparable sub-images of recognized objects \parencite[such as an ensemble space of human pose to further operationalize Aby Warburg's Mnemosyne, cf.][]{warburg_bilderatlas_2008,impett_pose_2016}. 
A scene in a film or a recorded theater play could be represented by the concatenation of a visual compression ensemble, a visual embedding, an audio compression ensemble, and a language model embedding of the spoken dialogue \parencite[e.g.][]{devlin_bert_2019-2}. The full apparatus of art history could further be combined with the presented approach, integrating our systematic study of visual aspects with socio-cultural contexts, as covered in literature and recorded in structured databases or knowledge graphs \parencite[cf.][]{schich_revealing_2010}.

Seen from yet another angle, we constitute three kinds of vector spaces: the multidimensional ensemble space of compression ratios, the decorrelated multidimensional space of associated PCA components, and a reduced two-dimensional UMAP space providing a proxy topography. Together these spaces may bring to mind the latent embedding spaces of deep machine learning, where the explanation of implicit dimensions remains a challenge.
Our three spaces can also be understood as subspaces of more general cultural meaning spaces. In the sense of Cassirer's ``most general reference framework" they can be seen as ``spaces of geometric intuition" \parencite[][]{cassirer_philosophie_2010,cassirer_symbolproblem_1927,schich_cultural_2019}, belonging to the realm of what art historians later called ``iconologic" aspects of visual art, complementing associated contextual information, including ``iconographic", i.e. written aspects \parencite[cf.][]{panofsky_studies_1939}. 
The three ensemble spaces are further in line with the cognitive theory of Gärdenfors \parencite[cf.][]{gardenfors_conceptual_2000,gardenfors_geometry_2014}, where conceptual spaces are based on a set of quality dimensions, and representations are rooted in topological and geometrical notions.
Finally, our approach resonates with the notion of information space \parencite[cf.][]{eigen_strange_2013}, where movement in space corresponds to a change in meaning. Given large sets of cultural products, we assume that a negotiation and integration of these various concepts of space may lead to further advances in the study of quantitative aesthetics.

Very recently, submitted concurrently, \textcite[][]{murphy_distributed_2022} have proposed a similar approach to reveal explanatory structure of complex systems centered around the concept of a ``distributed information bottleneck", including an analysis of the painting of Mona Lisa as one of their examples. While their model may seem more general at first glance, applied to several types of complex systems, our approach has a wider scope in another sense. 
Our compression ensemble approach aims to understand visual artifacts implying an algorithmic process in the broadest sense. 
This algorithmic process may include individual, collective and external cognition (for example, using pencil and paper to think), where the bandwidths of the ``distributed information bottleneck" are not necessarily a limiting factor.
Therefore, we argue that it is realistic to consider transformations which not only decrease but add information preceding compression, as we do in our proposed framework. 
An intuitive example of such a broadening of bandwidth using external cognition would be the transformation of large amounts of urban features into a city map, consecutively compressed via abstraction to postcard size.

\section{Conclusion} %

In summary, throughout this paper we have shown here the utility of the ensemble approach, which provides a vector-based (and therefore fast) algorithmic distance metric. 
While previous research in computational aesthetics and psychology has engaged in looking for a single metric of visual complexity, we argue that it may be useful to instead use an array of estimates that captures complimentary aspects of complexity.
While our proposal is particularly suited for the analysis of visual media, this approach holds broader promise as a new framework for the quantification of aesthetic, linguistic and cultural complexity.

\FloatBarrier
\section{Materials and methods}

\subsection{Constructing a vector space of algorithmic distance}

As discussed in the Introduction, compression as such has been used to estimate visual and aesthetic complexity before. In some applications, it has also included combination with limited visual transformations \parencite[][]{bagrov_multiscale_2020,mccormack_complexity_2022,lakhal_beauty_2020,machado_computerized_2015,fernandez-lozano_visual_2019}. However, the fairly large number of transformations is key to our approach, with the following rationale.
Consider two algorithmically similar uncompressed images $A$ and $B$, for example two versions of the same famous view of Rouen cathedral by Claude Monet (of which the artist painted more than 30 in 1892-1893). These two images will yield similar compressed sizes for the same compression algorithm because the ``algorithm" that generated them (being a function of Monet's perspective, style, and execution) is similar. Another artwork $C$, e.g. a late, abstract work by Piet Mondrian will, due its lack of detail, likely have a much smaller compression size. However it is entirely conceivable that a work $D$ that is stylistically very different to Monet's Rouen cathedral, e.g. a surrealist painting by Salvador Dali, might {\em by chance} have a very similar compression size. The ``algorithms" used by Monet and Dali differ greatly, and an equal compression size does not imply that they are of equal algorithmic complexity either, as the efficiency of the compression algorithm itself will differ depending on the detailed characteristics of the images. 
However, now consider an image transformation $T$ (e.g. Gaussian blur), which we apply to the uncompressed versions of our four images $A$, $B$, $C$, and $D$ before compressing them. The compressed sizes of $T(A)$ and $T(B)$ are still likely to be very similar, as the algorithms that generated the original images are very similar, and the transformation and compression algorithms are identical. $T(C)$ is very likely to still be very different to $T(A)$ and $T(B)$. While the compressed size of $D$ was similar to $A$ and $B$ by chance, it is much more unlikely that $T(D)$ is also similar to $T(A)$ and $T(B)$, as the interaction between the transformation $T$ and the generative algorithm of $D$ would have to change the compressibility in the same way as the interaction of $T$ and $A$/$B$. Put more intuitively, a Gaussian blur is very likely to affect the compressibility of a Monet very differently from the compressibility of a Dali. Thus, more generally speaking, two images with similar compressed sizes are much less likely to still yield similar compressed sizes by chance after a transformation {\em unless} they are algorithmically similar to start with in which case the combined algorithms of generation and transformation (and their interaction with the compression algorithm) remain similar. 
If we now consider the application of $N$ different transformations of an uncompressed image $A$, each applied before a subsequent compression, the compressed sizes (including of the untransformed image) $c(A)$, $c(T_1(A))$, $c(T_2(A))$, ... $c(T_N(A))$ form a vector ${\bf v}(A)$ of length $N+1$. It follows from the above argument about coincidental proximity that it becomes increasingly unlikely for two algorithmically dissimilar images to remain close together as $N$ increases. Thus the resulting vector space of compressed sizes provides an indication of algorithmic distance between images.

\subsection{Data processing and limitations}

In practice, we use normalized compression lengths. The compression size of the original image without transformations is divided by the size of the original bitmap image. Compressions of transformations are divided by the size of the original compression. In most applications discussed in this paper, it also makes sense to rescale the vector space components (we use z-scoring), to put the compression ratios and the additional statistics on a comparable scale.  

The statistical transformations include the following. We use both the LAB (``M2") and RGB space (``M3") based measures of colorfulness from \textcite[][]{hasler_measuring_2003}; standardize images by quantizing down to 200 colors and record statistics of contrast (range and standard deviation of the lightness channel values in LAB space), color distribution mean, median, max, standard deviation and entropy (which all provide insight into color complexity). We also attempt to estimate composition regularity, first as entropy of the angles of composition lines (based on the Hough transform applied on Canny filtered, i.e. edge-detected images). We also estimate fractal or Hausdorff dimension on bilevel-quantized versions of the images, using both a small and large window size, and on a Canny-filtered image \parencite[cf.][]{gneiting_estimators_2012}.

Both the Historical and Contemporary Hic et Nunc dataset are preprocessed the same way, downscaling images to 160000 pixel bitmaps (400x400 in the case of a perfect square) while retaining aspect ratio. Smaller images up to 50\% of that size are allowed (but not upscaled), smaller images are discarded. Another option would be to resize all images to identical squares, but that would distort the composition of wide or tall artworks. The aspect differences, size differences resulting from integer division of the 160000 and the inclusion of smaller images, are all controlled for in the next step. The assigned file size of a compressed image (or its transformation) is actually the mean of two compressions, of the original and its 90 degree rotation. The compression ratios are calculated in terms of the respective downscaled bitmaps. Furthermore, one of our visual transformations is the Fast Fourier Transform; given its square-shaped output components, the transform is applied twice, on the original and its rotation, and the resulting components are also additionally rotated for compression.

This approach to homogenizing the images is far from perfect, as the size of the originals that these photographs and scans represent may well range from the size of a postcard to that of an altar piece. Not only that, but the latter may well be represented by a lower resolution image than the former, with better or worse color grading, etc. The dataset contains sparse metadata on original size, but we have no way to systematically quantify this issue at scale, and remains a limitation of the current study. However, in a sense, our approach is not very different from making art historical inferences by going through and looking at large visual resource collections, much like students of art history examining art historical survey literature, or an art connoisseur training their eye using a large comprehensive 35mm slide collection of a library of photos, which historically served exactly this very purpose.

Since we are interested in making comparisons over time, the Historical dataset was also filtered for items with an identifiable creation date. We carried out some preprocessing of the date metadata, retrieving four-digit years from descriptions that included them. However, much of earlier art is tagged with heterogeneous and approximate descriptions such as ``early XVI century". Discarding these made the earlier end of this dataset even smaller, which is why we limit some analyses to the 19-20th century.
The Historical dataset is also likely biased in a number of ways. It features primarily Western art, most of the data is concentrated in the 20th century, the metadata quality varies and is of unidentifiable origin, the sampling mechanisms are unknown but likely biased by archival and selection practices of the various museums and collections these reproductions originate and the websites that house them.

In general, in the case of art collections or databases like Wikiart and art500k, it is important to be clear that these consist of small, curated, often biased samples of art of some place and period --- and as such, they represent the historiography of art first and the actual history of art second \parencite[see also][]{lee_dissecting_2020}. This is not unlike the case of linguistic corpora, which also consist of small curated samples of utterances (in diachronic corpora, often in the form of newspaper articles or books) from a much larger population of all utterances produced by all speakers of a given language over some period of time. In short, it is important to acknowledge that when we make claims here about the history or dynamics of visual art, we are only referring to information derived from the sample --- but we make the assumption that the sample is reasonably representative and as such informative of the population of Western art in the time periods we cover.
This means that the figures depicting historical changes may look different if more data would be available. However, this is not a weakness of our approach, but an opportunity to use it for the study of data set bias. %
Indeed we are confident that future research will confirm our results in principle, while making headway by enhancing the approach, with larger more complete datasets as they become available.

\subsection{Adjustment of temporal resemblance}

As an important technical detail, we need to adjust the distances in the Temporal Resemblance model reported in the Results section, due to two biasing factors in the Historical dataset: the boundedness of the dataset (works in the last years have a higher likelihood of having neighbors in the past and vice versa) and its imbalance (much more data in some decades than others). The adjustment works as follows. We calculate temporal distances for all works between 1800-1990. We limit ourselves to this period, where the amount of works per unit of time is more consistent, compared to other parts of the corpus where there is less data but also more variation between years in terms of data points. When finding nearest neighbors, the entire dataset is still taken into account, e.g. a work from 1801 can theoretically have one of its nearest neighbor in 1400 or 2018. We then fit a generalized additive regression model, predicting distance by year. The residuals from that model, still on the scale of years, approximate temporal resemblance given the shape and bounds of the data.

\FloatBarrier
\section*{Author contributions, acknowledgments and funding} 

A.~K., M.~S., and S.~E.~A. co-designed the research. A.~K. also prepared data, designed and performed data analysis, co-wrote the text, and created the figures. M.~S. also co-designed the data analysis and figures, performed preliminary analysis, co-wrote the text, and provided conceptual guidance. S.~E.~A. designed the compression ensemble algorithm as documented in a preliminary manuscript, performed preliminary analysis, co-wrote the text and provided conceptual guidance for the analysis and figures. M.~C.~S. and T.~O. contributed to the design of the analysis and figures, and performed data mining.
A.~K., M.~C.~S., T.~O., and M.~S. are supported by the CUDAN ERA Chair project, funded through the European Union’s Horizon 2020 research and innovation program (Grant No. 810961).
S.~E.~A. was funded by the Royal Society as a University Research Fellow during some of the time of his work on this.
We would like to thank Dr. Mikhail Tamm for helpful discussions.
Thumbnail previews of artworks depicted for informative purposes as fair use.

\begingroup
\setlength{\emergencystretch}{8em}
\printbibliography
\endgroup
\FloatBarrier
\newpage
\section*{Supplementary materials} 
\renewcommand\thefigure{S\arabic{figure}}    
\setcounter{figure}{0}  
\renewcommand\thetable{S\arabic{table}}    
\setcounter{table}{0}

\subsection*{List of transformations in the compression ensemble}

Table \ref{tab_trans} lays out all the transformations used in the version of the compression ensemble used in this paper. The flood fill color is determined by finding a primary color that is least frequent in the distribution of pixel colors of the input image. 

\begin{longtable}{
    @{}
    p{0.02\textwidth}p{0.22\textwidth}p{0.06\textwidth}p{0.04\textwidth}p{0.53\textwidth}}
\caption{List of transformations in the compression ensemble used in our analyses. 
The Type values stand for the following. C: compression without transformation, CT: compression of transformed image, S: statistical transformation without involving compression as such. The \textsc{jpeg} algorithm is used in two modes, with the quality parameter at 100 and at 0. Sizes refer to image dimensions (after initial normalization) --- values smaller than one indicate fractional size, e.g. 0.4 stands for width and height reduce to 40\% of the normalized size, before applying transformations.}
\label{tab_trans}
\\  \hline
Type & \vspace{1mm} Feature & \makecell[tl]{Comp-\\ ressions} & Sizes & Description \\ 
  \hline
   C & compress & gif, png, jpeg100, jpeg0 & 1, .4, .2, .1 & Compression of the original, of various resizes and algorithms \\ 
  CT & blur10 & gif, png & 1, .4 & Blur filter (radius, sigma=10) \\ 
  CT & blur30 & gif & 1 & Blur with (radius, sigma=30) \\ 
  CT & colors\_grayscale & gif & 1, .4 & Grayscale filter \\ 
  CT & colors\_quantize\_bw & gif & 1, .4 & Black and white filter \\ 
  CT & colors\_quantize\_bw\_dither & gif & 1, .4 & Black and white filter with dithering \\ 
  CT & colors\_quantize3 & gif & 1, .4 & Color quantization (n=3) \\ 
  CT & colors\_quantize5 & gif & 1 & Color quantization (n=5) \\ 
  CT & colors\_acos & gif & 1, .4 & Color values transformation, arc cosine \\ 
  CT & colors\_p10 & gif & 1, .4 & Color values transformation, power of 10 \\ 
  CT & colors\_sqrt & gif & 1 & Color values transformation, square root \\ 
  CT & colors\_round & gif & 1 & Color values transformation, rounding \\ 
  CT & colors\_brightness & gif & 1 & Color brightness increased by 300$\backslash$\%  \\ 
  CT & colors\_saturate & gif & 1 & Color saturation increased by 5000$\backslash$\% \\ 
  CT & color\_chroma\_divide & gif & 1 & Color values divided by chroma channel values \\ 
  CT & colors\_roundchroma & gif & 1 & Color transformation, rounding chroma channel values \\ 
  CT & color\_luminance\_divide & gif & 1 & Color transformation, dividing color by the luminance channel \\ 
  CT & color\_luminance\_lighten & gif & 1 & Color transformation, adding the luminance channel as linear light effect \\ 
  CT & color\_darken\_intensity & gif & 1 & Color values transformation, pairwise comparison to negative + intensity filter \\ 
  CT & colors\_add2 & gif & 1 & Color values transformation, add-composite of the image with itself \\ 
  CT & lines\_bw\_canny & gif & 1, .4 & Line (edge) detection, Canny filter \\ 
  CT & lines\_cartoon & gif & 1, .4 & Line detection, multiply-composite of Canny and blurred image \\ 
  CT & lines\_division\_gray & gif, png & 1, .4 & Line detection via division with blurred grayscale of itself \\ 
  CT & lines\_edge5\_gray & gif, png & 1, .4 & Line detection based on grayscaled image (radius=5) \\ 
  CT & lines\_edge10\_gray & gif & 1, .4 & Line detection based on grayscaled image (radius=10) \\ 
  CT & lines\_edge1\_color & gif & 1, .4 & Line detection based on full color image (radius=1) \\ 
  CT & lines\_edge2\_color & gif & 1 & Line detection based on full color image (radius=2) \\ 
  CT & lines\_hough40 & gif & 1 & Line detection, Hough filter (width=40, height=40, threshold=20) \\ 
  CT & lines\_hough50 & gif & 1, .4 & Line detection, Hough filter (width=50, height=50, threshold=70) \\ 
  CT & lines\_color\_comp & gif & 1 & Line detection via comparison to despecled original \\ 
  CT & lines\_color\_conv & gif & 1 & Line detection, Sobel convolution-based \\ 
  CT & lines\_edge\_lat & gif & 1 & Line detection, Local Adaptive Thresholding \\ 
  CT & emboss\_gray4 & gif & 1, .4 & Emboss filter on grayscaled image (radius=4, sigma=1) \\ 
  CT & emboss\_col1 & gif & 1 & Emboss filter on full color image (radius=1, sigma=0.1) \\ 
  CT & emboss\_conv\_grayd & gif & 1 & Emboss-like effect on grayscale image via convolution with DoG kernel \\ 
  CT & emboss\_conv\_grayp & gif & 1 & Emboss-like effect on grayscale image via convolution with Prewitt kernel \\ 
  CT & emboss\_modulate & gif & 1 & Emboss-like effect via modulate-compositing the image with itself and despecling \\ 
  CT & morph\_pixelate10 & gif & 1, .4 & Pixelation via rescaling down 10x and back to original size \\ 
  CT & morph\_pixelate20 & gif & 1, .4 & Pixelation via rescaling down 20x and back to original size \\ 
  CT & morph\_add3\_pixelate & gif & 1 & Pixelation via add-compositing the image with itself 3x \\ 
  CT & morph\_despecle10 & gif & 1 & Despecle filter, 10x \\ 
  CT & morph\_oilpaint & gif & 1 & Oil painting effect \\ 
  CT & morph\_squares & gif & 1 & Square-shaped dilation effect \\ 
  CT & flood\_centre & gif & 1, .4 & Flood fill image with single color, from centre \\ 
  CT & flood\_hole & gif & 1, .4 & Flood fill image with single color, filled circle \\ 
  CT & flood\_corners & gif & 1 & Flood fill image with single color, from corners \\ 
  CT & flood\_thirds & gif & 1 & Flood fill image with single color, from intersections of thirds \\ 
  CT & fx\_deskew\_zoom & gif & 1, .4 & Auto-deskew and crop to center \\ 
  CT & fx\_implode & gif & 1 & Gravity well visual effect at center \\ 
  CT & fx\_noise & gif & 1 & Multiplicative noise effect \\ 
  CT & fx\_scramble & gif & 1 & Randomly scramble positions of all pixels \\ 
  CT & fx\_stripes & gif & 1 & Quantize to 20 colors and order pixels by color frequency \\ 
  CT & fft1 & gif & 1 & Fast Fourier Transform of image, magnitude output \\ 
  CT & fft2 & gif & 1 & Fast Fourier Transform of image, phase output \\ 
  CT & fft1\_blur10 & gif & 1 & Fast Fourier Transform of blurred image, magnitude output \\ 
  CT & fft2\_blur10 & gif & 1 & Fast Fourier Transform of blurred image, phase output \\ 
  CT & fft1\_colors\_quantize3 & gif & 1 & Fast Fourier Transform of colors\_quantize3, magnitude output \\ 
  CT & fft2\_colors\_quantize3 & gif & 1 & Fast Fourier Transform of colors\_quantize3, phase output \\ 
  CT & fft1\_lines\_division\_gray & gif & 1 & Fast Fourier Transform of lines\_division\_gray, magnitude output \\ 
  CT & fft2\_lines\_division\_gray & gif & 1 & Fast Fourier Transform of lines\_division\_gray, phase output \\ 
  S & fractaldim1 & NA & 1 & Fractal dimension estimate (window=m/10, step=m/40, where m is the mean of x and y dimensions) \\ 
  S & fractaldim2 & NA & 1 & Fractal dimension estimate (window=m/3, step=m/4) \\ 
  S & fractaldim3 & NA & 1 & Fractal dimension estimate of lines\_bw\_canny (window=m/5, step=m/6) \\ 
  S & stats\_angleentropy & NA & 1 & Entropy of angles in lines\_hough40  \\ 
  S & stats\_colfreq\_entropy & NA & 1 & Color frequency distribution: entropy \\ 
  S & stats\_colfreq\_max & NA & 1 & Color frequency distribution: maximum \\ 
  S & stats\_colfreq\_mean & NA & 1 & Color frequency distribution: mean \\ 
  S & stats\_colfreq\_median & NA & 1 & Color frequency distribution: median \\ 
  S & stats\_colfreq\_sd & NA & 1 & Color frequency distribution: standard deviation \\ 
  S & stats\_colorfulness\_lab & NA & 1 & Color complexity estimate M2, LAB space  \\ 
  S & stats\_colorfulness\_rgb & NA & 1 & Color complexity M3, RGB space  \\ 
  S & stats\_contrastrange & NA & 1 & Contrastiveness estimate (luminosity values absolute range) \\ 
  S & stats\_contrastsd & NA & 1 & Contrastiveness estimate (luminosity values standard deviation \\ 
   \hline
\end{longtable}

\subsection*{Compression ensemble pipeline technical details} 

This section gives a step-by-step overview of the pipeline used in this work implementing the compression ensemble approach. The pipeline was implemented in R (version 4.1.2). The visual transformations and export into various compression formats is handled by the magick R package (2.7.3), which is a wrapper for the Imagemagick cross-platform software suite (6.9.12.3). 

The compressed file sizes are recorded as the mean of two sizes, that of compression of the image in the original orientation, and after 90 degree rotation (as these values usually differ slightly due to the way image compression works). For the Fast Fourier Transform, and additional rotation step is is applied both before doing FFT (as FFT in Imagemagick uses the width of the original image for the dimensionality of the square-shaped outputs).

We recommend carrying out the image compression steps in memory, as saving all the files to disk would be computationally inefficient. Parallelization of the pipeline is recommended if the dataset is large. Note that due to the architecture of Imagemagick, the file sizes of image outputs in different formats may vary slightly between operating systems. As long as all images in a given dataset go through the pipeline on the same system, this should not be a problem though. We carried out all processing on an Nvidia DGX Station A100 running Red Hat Enterprise Linux (AS release 4) hosting a Docker running Ubuntu 20.04.3.

The vector of compression ratios and statistical transformations is calculated as follows.
\vspace{-\topsep}
\begin{itemize}
 \setlength{\parskip}{0pt}
  \setlength{\itemsep}{0pt plus 1pt}
  \item Import image file, convert to RGB colorspace, downsize so the total number of pixels $n$ is (as close as possible to) 160000. Smaller images are not upsized. If the original image is smaller than 50\% of that, stop processing.
  \item Record the RGB bitmap file size of this normalized image as $f$
  \item Compress the image using \textsc{gif} compression, record the ratio of the size of this file to $f$ as $b_{gif}$. This is the baseline value that all transformations compressed using \textsc{gif} will be compared to. 
  \item Also compress the image using other compression algorithms (\textsc{png, jpeg} --- we do the latter both with quality parameter 0 and 100), record the ratio, these file sizes divided by $f$. Record the \textsc{png} size as $b_{png}$.
  \item Downscale the image to 40, 20 and 10\% of $n$, record compression ratio to $f$ with the different algorithms.
  \item Transform the image using all the visual transformations $CT$ (see list above), compress using \textsc{gif}, record the ratio $c$ by dividing the compressed file size with $b_{gif}$
  \item Also transform the image using a subset of the visual transformations, compress using \textsc{png}, record the ratio $c$ by dividing the compressed file size with $b_{png}$
  \item Carry out the statistical transformations $S$ (see list above).
  \item Concatenate the compression ratios and statistical transformation values into a vector.
\end{itemize}\vspace{-\topsep}
  
Note that the compression ensemble approach is general and the specific number of transformations is unimportant. As demonstrated in the Evaluation section in the main text, and handful transformations may be enough for some tasks. Here, we opted to implement a fairly large array of transformations and additional compressions with different sizes and algorithms, in order to explore the transformation space in a relatively comprehensive manner. If, unlike here, computation time is important, a smaller set of transformations is likely a better choice.

\FloatBarrier
\newpage
\subsection*{Transformations mapped onto UMAP}

\begin{figure}[htb]
	\noindent
	\includegraphics[width=\columnwidth]{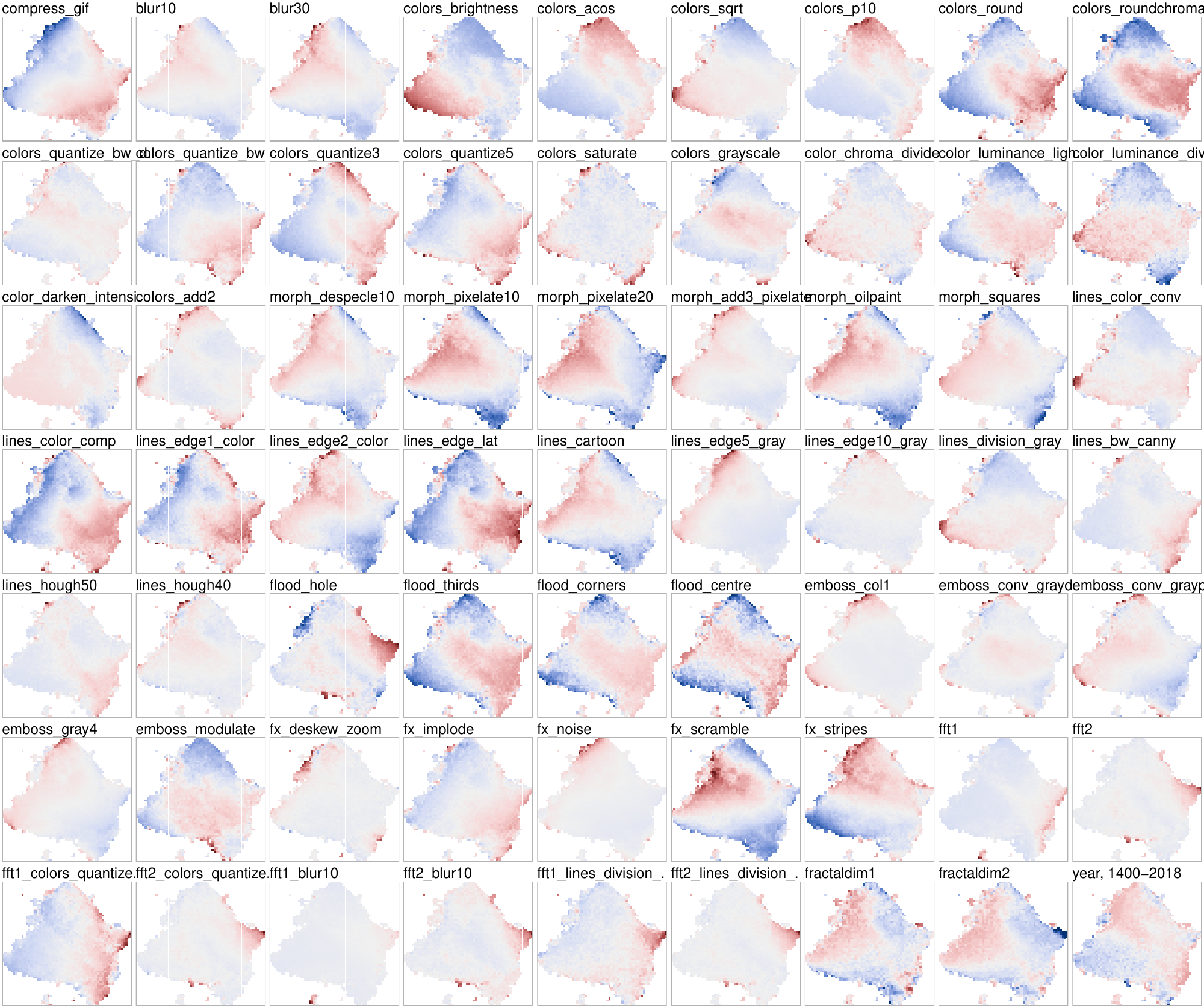}
	\caption{The transformations illustrated in Figure \ref{fig_illustrate} mapped onto the UMAP projection of Figure \ref{fig_bigpca}, here as a heatmap (for better visibility), where each cell represents the mean value of the points in a given area of the UMAP. Blue stands for low, gray for mean and red for high values in a given compression ratio or variable. Date of creation is added as an additional map to the bottom right.
	}\label{fig_sup_smallmaps}
\end{figure}

Figure \ref{fig_sup_smallmaps} illustrates individual transformations and how they constitute the UMAP shown in Figure \ref{fig_bigpca}.

\FloatBarrier
\newpage
\subsection*{Style period confusion matrix}

\begin{figure}[htb]
	\noindent
	\includegraphics[width=\columnwidth]{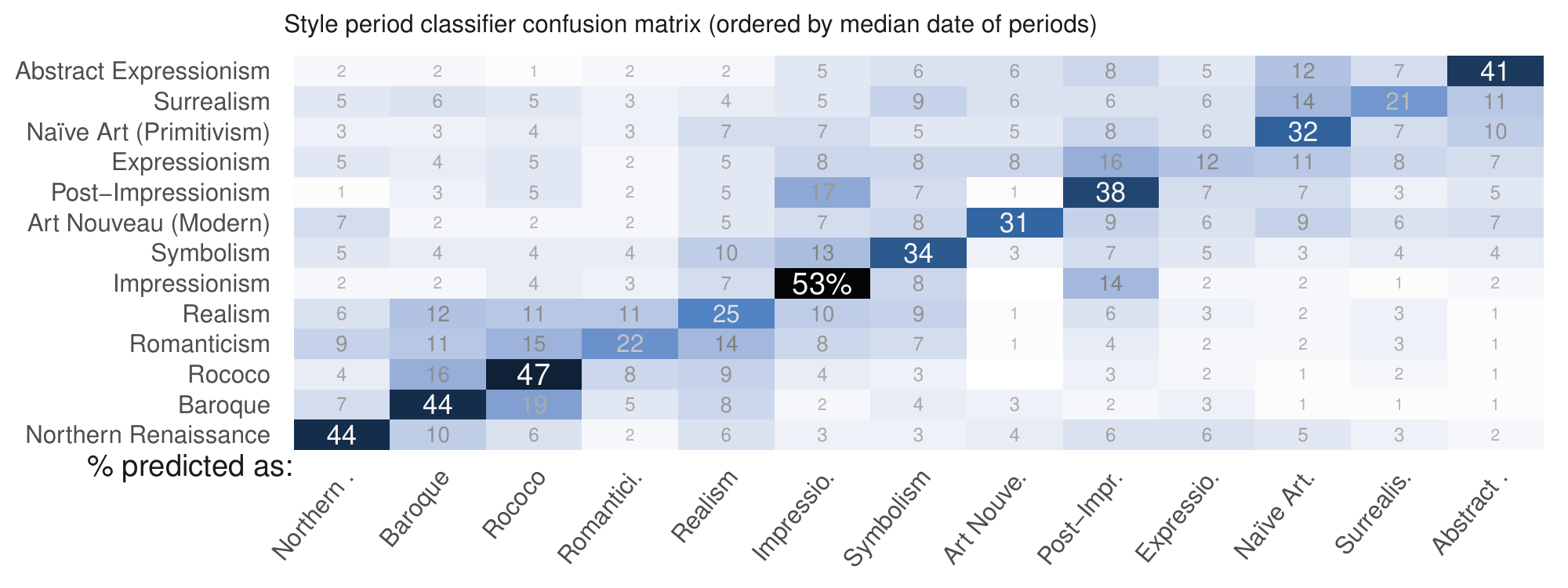}
	\caption{Confusion matrix of the style period classifier. Each row represents the percentage a given style gets predicted as another style; the diagonal stands for correct predictions. The values are from a LDA classifier using 1000 training examples and the full set of transformations; the values are means of 1000 re-runs. This figure illustrates the extent a simple classifier manages to correctly predict style based on the compression ensemble vectors, and which styles are more often confused than others.
	}\label{fig_sup_stylecm}
\end{figure}

Figure \ref{fig_sup_stylecm} supplements the discussion on the art classifier in the Evaluation section in the main text (see also Figure \ref{fig_evals}).

\FloatBarrier
\subsection*{Artworks in temporal resemblance}

\begin{figure}[thb]
	\noindent
	\includegraphics[width=\columnwidth]{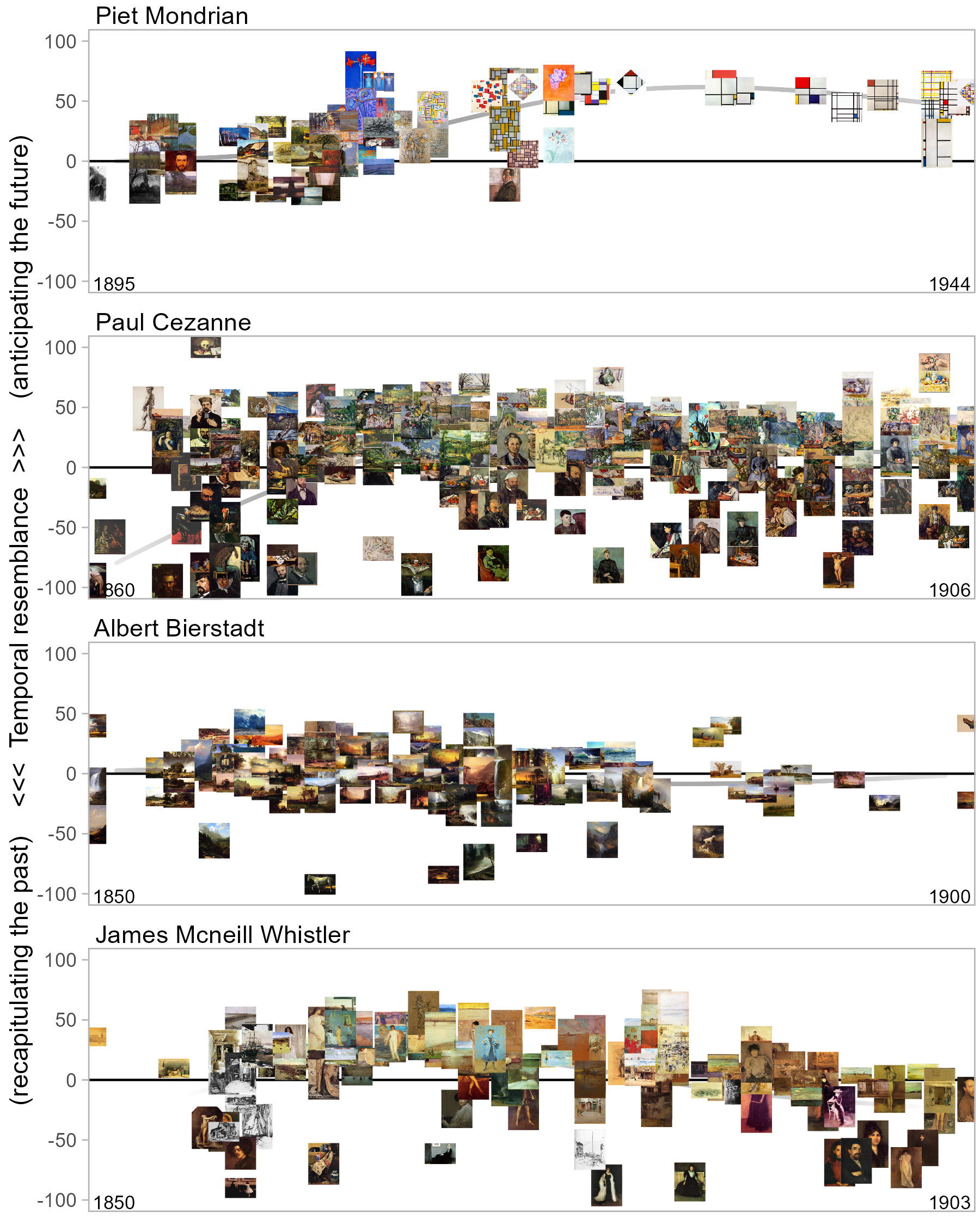}
	\caption{All artworks on the main panels of Figure \ref{fig_temporal}, centered on their coordinates on the vertical and horizontal axes.
	}\label{fig_sup_temporal}
\end{figure}

Figure \ref{fig_sup_temporal} supplements Figure \ref{fig_temporal}, displaying all the artworks as larger thumbnails.

\FloatBarrier
\subsection*{Results of pairwise vector additions}

\begin{figure}[thb]
	\noindent
	\includegraphics[width=\columnwidth]{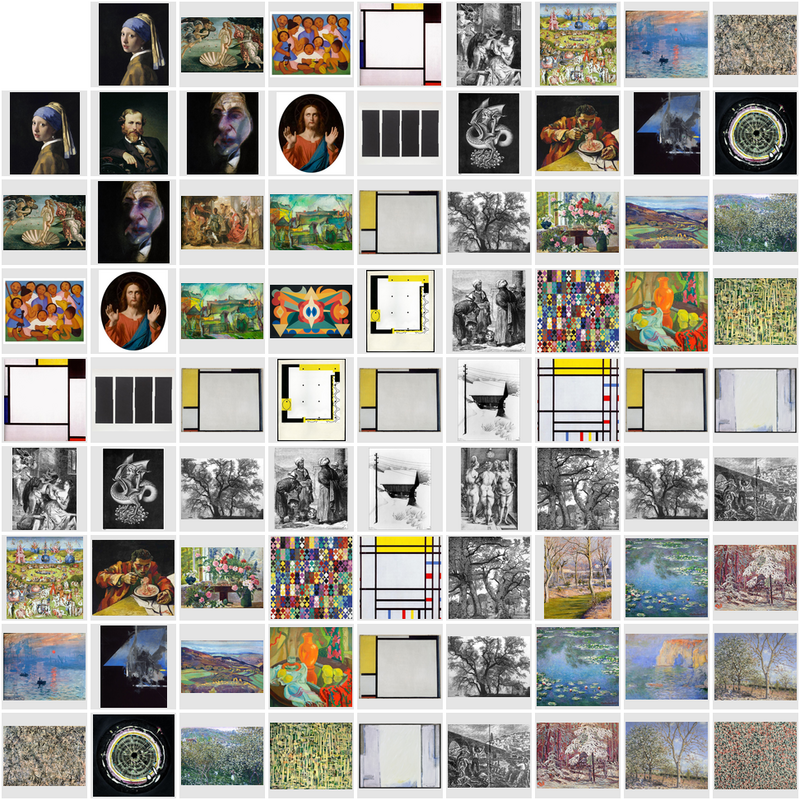}
	\caption{This figure exemplifies the results of vector operations discussed in the main text as a form of latent space navigation, and supplements \ref{fig_bigpca}. Each cell is a result of the following operation: take the compression ensemble vectors of the first image in a given row and column, sum them element-wise, find the cosine-nearest neighbor in the ensemble for that new vector. This figure shows how vector addition can be used to navigate the space of aesthetic complexity.
	}\label{fig_sup_matsum}
\end{figure}

\end{document}